\documentclass{llncs}

\usepackage{multicol}

\usepackage{amsfonts}
\usepackage{amsmath} 
\usepackage{amssymb}  
\usepackage{graphicx}
\usepackage{graphics} 
\usepackage{epsfig} 
\usepackage{mathptmx} 
\usepackage{times} 
\usepackage{url}
\usepackage{rotating}
\usepackage{subfigure}
\usepackage{multicol}
\usepackage{algorithm}              
\usepackage{algorithmic}
\usepackage{caption}            
\usepackage{multirow}
\usepackage{hyperref}
\newcommand{\stt}[1]{{\small\texttt{#1}}}

\newenvironment{small_ind_s_itemize}{\begin{list}{$\bullet$}
{\setlength{\rightmargin}{0em}
\setlength{\leftmargin}{1em}
\setlength{\itemsep}{0em}
\setlength{\topsep}{0em}
\setlength{\parsep}{0em}}}{\end{list}}

\newcounter{ctr}

\title{\bf Combining Answer Set Programming and POMDPs for Knowledge
  Representation and Reasoning on Mobile Robots}

\author{Shiqi Zhang and Mohan Sridharan}
\institute{
  Department of Computer Science\\ 
  Texas Tech University, USA \\
 \email{shiqi.zhang6@gmail.com, mohan.sridharan@ttu.edu}
}

\begin{document}
\label{firstpage}
\maketitle

\begin{abstract}
  For widespread deployment in domains characterized by partial
  observability, non-deterministic actions and unforeseen changes,
  robots need to adapt sensing, processing and interaction with humans
  to the tasks at hand. While robots typically cannot process all
  sensor inputs or operate without substantial domain knowledge, it is
  a challenge to provide accurate domain knowledge and humans may not
  have the time and expertise to provide elaborate and accurate
  feedback. The architecture described in this paper combines
  declarative programming and probabilistic reasoning to address these
  challenges, enabling robots to: (a) represent and reason with
  incomplete domain knowledge, resolving ambiguities and revising
  existing knowledge using sensor inputs and minimal human feedback;
  and (b) probabilistically model the uncertainty in sensor input
  processing and navigation. Specifically, Answer Set Programming
  (ASP), a declarative programming paradigm, is combined with
  hierarchical partially observable Markov decision processes
  (POMDPs), using domain knowledge to revise probabilistic beliefs,
  and using positive and negative observations for early termination
  of tasks that can no longer be pursued. All algorithms are evaluated
  in simulation and on mobile robots locating target objects in indoor
  domains.
\end{abstract}

\vspace{-2em}
\section{Introduction}
\vspace{-0.5em}
\label{sec:intro}
Mobile robots are increasingly being deployed to collaborate with
humans in domains such as search and rescue, reconnaissance, and
surveillance.  These domains characterized by partial observability,
non-determinism and unforeseen changes frequently make it difficult
for robots to process all sensor inputs or operate without substantial
domain knowledge. At the same time, it is a challenge to provide
complete domain knowledge in advance, and humans are unlikely to have
the time and expertise to provide elaborate and accurate feedback.
Widespread deployment of robots in our homes, offices and other
complex real-world domains thus poses formidable knowledge
representation and reasoning (KRR) challenges---robots need to: (a)
represent, revise and reason with incomplete knowledge; (b)
automatically adapt sensing and acting to the task at hand; and (c)
learn from unreliable, high-level human feedback.

Although there is a rich body of research on knowledge representation
and reasoning, the research community is fragmented. For instance,
declarative programming paradigms provide commonsense reasoning
capabilities for robotics but do not support probabilistic modeling of
uncertainty, which is essential in robot application domains. In
parallel, sophisticated algorithms based on probabilistic graphical
models are being designed to model the uncertainty in sensing and
navigation on robots, but it is difficult to use these algorithms for
commonsense reasoning. Furthermore, algorithms developed to combine
logical and probabilistic reasoning do not provide the desired
expressiveness for commonsense reasoning capabilities such as
non-monotonic reasoning and default reasoning.  Our previous work
described an architecture that combined the appealing KRR capabilities
of \emph{Answer Set Programming} (ASP), a declarative programming
paradigm, with the probabilistic uncertainty modeling capabilities of
\emph{partially observable Markov decision processes}
(POMDPs)~\cite{zhang:icdl12}. This paper significantly extends the
architecture by making the following contributions:
\begin{small_ind_s_itemize}
\item Richer representation of domain knowledge in ASP, incrementally
  revising the knowledge base (KB) with information obtained from
  sensor inputs and human feedback.
\item Principled generation of prior beliefs from the KB, which are
  merged with POMDP beliefs using Bayesian updates to adapt sensing
  and acting to the tasks at hand.
\item Modeling and learning from positive and negative observations,
  identifying situations in which the current task should no longer be
  pursued.
\end{small_ind_s_itemize}
All algorithms are evaluated in simulated and physical robots visually
localizing target objects in indoor domains. Section~\ref{sec:relatedWork}
discusses limitations of existing work to motivate algorithms described in
Section~\ref{sec:algorithm}. Section~\ref{sec:expres} presents experimental
results, followed by conclusions in Section~\ref{sec:conclusions}.

\vspace{-1em}
\section{Related Work}
\vspace{-0.5em}
\label{sec:relatedWork}
Algorithms based on probabilistic graphical models such as POMDPs have
been used to model the uncertainty in real-world sensing and
navigation, enabling the use of robots in offices and
hospitals~\cite{Gobelbecker:aaai11,Pineau:RAS03,Rosenthal:aaai11}.
Since the rapid increase in state space dimensions of such
formulations of complex problems make real-time operation difficult,
researchers have developed algorithms that decompose complex problems
into a hierarchy of simpler problems that are computationally
tractable~\cite{Pineau:RAS03,zhang:aamas12}. However, it is still
challenging to use POMDPs and other graphical models in large, complex
state-action spaces. Furthermore, these probabilistic algorithms do
not readily support representation of, and reasoning with, commonsense
domain knowledge.

There is a rich body of research on knowledge representation and
reasoning. Declarative programming paradigms such as ASP provide
appealing capabilities for non-monotonic logical reasoning and default
reasoning~\cite{Baral:book03,Gelfond:book08}. ASP has been used for
commonsense representation and reasoning for a robot
housekeeper~\cite{Erdem:ISR2012}, and for representing domain
knowledge learned through natural language
processing~\cite{Chen:JHRI2012}. However, ASP does not support
probabilistic modeling of the uncertainty in real-world sensing and
navigation.  Algorithms developed in recent years to combine logical
and probabilistic reasoning do not fully exploit the complementary
properties of logic programming and probabilistic models. For
instance, the \emph{switching planner} uses a POMDP or a DTPDDL
formulation to control a robot's behavior~\cite{Gobelbecker:aaai11},
but the threshold used to switch between the two formulations may
result in available information not being fully utilized. Commonsense
knowledge and semantic maps have also been used in motion and
task-level planning~\cite{Hanheide:ijcai11,Kaelbling:iros11}, but
accurate domain knowledge or extensive human supervision may not be
readily available.  Principled approaches for combining first-order
logic and probabilistic reasoning include Markov logic networks
(MLNs), which assign weights to logic
formulas~\cite{richardson2006markov}, and Bayesian Logic (BLOG), which
relaxes the unique name constraint of first-order probabilistic
languages to provide a compact representation of probability
distributions over outcomes with varying sets of
objects~\cite{Milch:bookchap07}. However, these algorithms do not
provide the desired expressiveness for capabilities such as
non-monotonic reasoning and default reasoning, which are important for
human-robot collaboration. Our prior work demonstrated the integration
of ASP with POMDPs on mobile robots~\cite{zhang:icdl12}, but did not
fully support capabilities such as incremental knowledge augmentation
and Bayesian belief merging. The architecture described in this paper
addresses these limitations.

\vspace{-1em}
\section{Proposed Architecture}
\begin{figure}[tbc] 
  \begin{center}
    \includegraphics[width=0.67\textwidth]{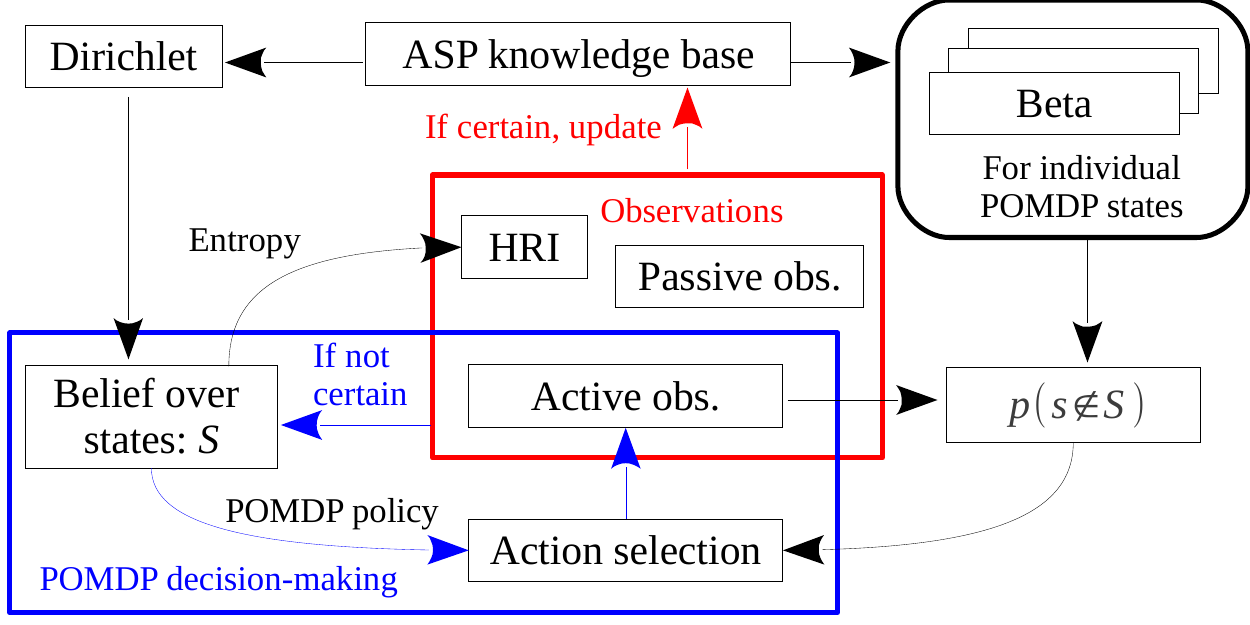}
    \vspace*{-1em}
    \caption{Overview of architecture integrating ASP and hierarchical
      POMDP.}
    \label{fig:overview}
    \vspace{-2.5em}
  \end{center}
\end{figure}

\vspace{-0.5em}
\label{sec:algorithm}
Figure~\ref{fig:overview} shows our control architecture. The ASP
knowledge base (KB) represents domain knowledge
(Section~\ref{sec:asp}), while POMDPs probabilistically model a subset
of the state space relevant to the task at hand
(Section~\ref{sec:pomdp}).  Logical inference in ASP is used to: (a)
compute prior beliefs relevant to the task, modeled as a
\emph{Dirichlet} distribution; and (b) identify eventualities not
being considered by the POMDP formulation based on \emph{Beta}
distributions.  The Dirichlet distribution supports Bayesian merging
with POMDP beliefs (Section~\ref{sec:initDistributions}), while Beta
distributions aids in the use of positive and negative observations
for early termination of tasks that cannot be accomplished
(Section~\ref{sec:nonexist}).

The robot uses a learned \emph{policy} to map merged POMDP beliefs to
action choices. Action execution causes the robot to move and process
images to obtain observations.  Observations are also obtained from
passive sensors (e.g., range finders) and high-level human feedback.
Observations made with high certainty are (proportionately) more
likely to be added into the KB, while other observations update POMDP
beliefs. Human feedback is solicited based on availability and need
(computed using the entropy of POMDP beliefs). Although this
architecture is illustrated below in the context of robots localizing
(i.e., computing the locations of) objects in indoor domains, the
integration of knowledge representation, non-monotonic logical
reasoning and probabilistic uncertainty modeling is applicable to many
domains.

\vspace{-1em}
\subsection{Knowledge representation and reasoning with ASP}
\vspace{-.5em}
\label{sec:asp}
Answer Set Programming is a declarative programming paradigm that can
represent recursive definitions, causal relations, and other
constructs that occur in non-mathematical domains and are difficult to
express in classical logic
formalisms~\cite{Baral:book03,Gelfond:book08}.  ASP's signature is a
tuple of sets: $\Sigma=\langle\mathcal{O}, \mathcal{F}, \mathcal{P},
\mathcal{V}\rangle$ that define names of objects, functions,
predicates and variables available for use. The signature is augmented
by \emph{sorts} (i.e., \emph{types}) such as: \stt{room/1},
\stt{object/1}, \stt{class/1}, and \stt{step/1} (for temporal
reasoning). An ASP program is a collection of statements describing
objects and relations between them. An \emph{answer set} is a set of
ground literals that represent beliefs of an agent associated with the
program. ASP introduces new connectives: \emph{default negation}
(negation by failure) and \emph{epistemic disjunction}, e.g., unlike
``{\tt $\lnot$ a}'' (classical negation), which implies that ``\emph{a
  is believed to be false}'', ``{\tt not a}'' only implies that
``\emph{a is not believed to be true}''; and unlike ``{\tt p
  $\lor\,\,\lnot$p}'' in propositional logic, ``{\tt p or $\lnot$p}''
is not a tautology.

\begin{figure}[tbc]
  \begin{center}
    \includegraphics[width=0.9\textwidth]{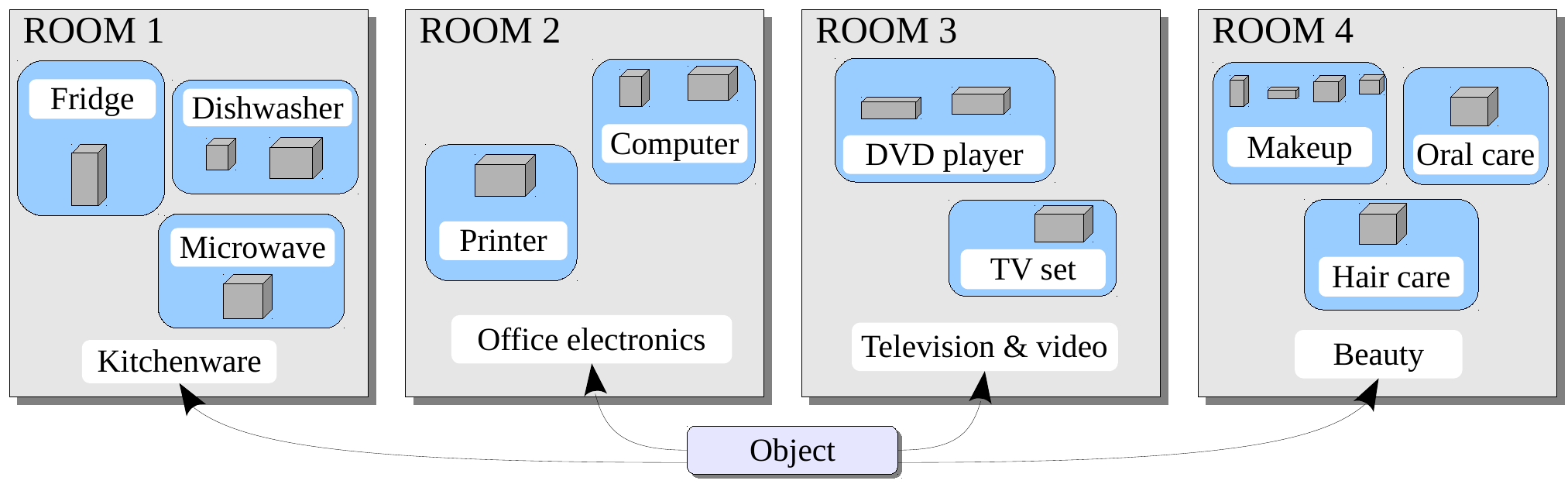}
    \vspace*{-1em}
    \caption{Pictorial representation of a hierarchy of objects.}
    \label{fig:example}
    \vspace*{-2em}
  \end{center}
\end{figure}

In the target localization domain, the robot learns a domain map and
semantic labels for rooms (e.g., ``kitchen'', ``office'').  Knowledge
about hierarchy of objects is learned incrementally from sensor
inputs, online repositories and human
feedback---Figure~\ref{fig:example} shows such a hierarchy.  This
hierarchy is revised using observations---specific object instances
(i.e., leaf nodes) can be added or removed and classes (leaf nodes'
parents are \emph{primary classes}) can be merged. The domain is thus
\emph{not} static.

The robot models and reasons about aspects of the domain that do not
change (\emph{statics}) and can change (\emph{fluents}), using
\emph{predicates} defined in terms of sorts of their arguments: (1)
\stt{is(object, class)} describes class membership of an object, e.g.,
\stt{is(printer1, printer)}; (2) \stt{subclass(class, class)} denotes
class hierarchy; (3) \stt{in(object, room)} describes the room
location of an object; (4) \stt{exists(class, room)}, implies that an
instance of a specific class exists in a specific room; and (5)
\stt{holds(fluent, step)} implies a fluent is true at a timestep.
Predicates are applied recursively when appropriate. The KB also
includes hand-coded \emph{reasoning} and \emph{inertial} rules such as
(assuming at least two rooms in the domain): 

\vspace*{-1em} 
{\small
  \begin{align*}
   (1)\,\, &\mathtt{holds(exists(C,R),I)}\,\,\leftarrow\,\, \mathtt{holds(in(O,R),I),~~is(O,C).}\\
   (2)\,\, &\mathtt{holds(exists(C1,R),I)}\,\,\leftarrow\,\,
    \mathtt{holds(exists(C2,R),I)},~~\mathtt{subclass(C2,C1).} \\
   (3)\,\, &\lnot\mathtt{holds(in(O,R2),I)}\,\,\leftarrow\,\, \mathtt{holds(in(O,R1),I),~~R1 != R2.}\\
   (4)\,\, &\mathtt{holds(in(O,R1),I+1)}\,\,\leftarrow\,\,
    \mathtt{holds(in(O,R1),I)},~~\mathtt{not}~~\mathtt{holds}\mathtt{(in(O,R2),I+1),~~R1
      != R2.}
  \end{align*} 
}The first rule implies that if object \stt{O} of class \stt{C} is in
room \stt{R}, it is inferred that an object of class \stt{C} exists in
\stt{R}; the second rule applies the first rule in the object
hierarchy. The last rule implies that if object \stt{O} is in room
\stt{R1}, it will continue to be there unless it is known to be
elsewhere.  As an example of non-monotonic reasoning, consider the
program: \stt{step(1..end).} \stt{is(printer1, printer).}
\stt{holds(in(printer1, lab), 1).} Now, consider adding a new
statement \stt{holds(in(printer1, office), 2)} about a printer's
location in the second time step. The answer sets obtained by
reasoning in ASP \emph{with} and \emph{without} this statement are
shown below (existing facts not repeated):

\vspace*{-1em}
{\small
  \begin{align*}
    \begin{array}{c|c}
      \underline{\texttt{With}} & \underline{\texttt{Without}} \\
      \mathtt{holds(in(printer1, lab), 2).}\,\, &\,\,\, \lnot\mathtt{holds(in(printer1, lab), 2).} \\
      \mathtt{holds(exists(printer, lab), 2).}\,\, &\,\,\,  \mathtt{holds(exists(printer, office), 2).}
    \end{array}
  \end{align*} 
}Addition of the new statement thus revises the outcome of the
previous inference step. Next, consider modeling the default:
``normally robots cannot climb stairs'' with humanoids encoded
elegantly as a \emph{weak exception}:

\vspace*{-1.3em}
{\small
  \begin{align*}
    & \lnot \mathtt{clmbstair(X)}\leftarrow\,\,\mathtt{robot(X),}~~~
    \mathtt{not}~~~\mathtt{ab(d}_{clmbstair}\mathtt{(X)).}\quad
    \textrm{\% Default rule}\\
    & \mathtt{robot(X)}\leftarrow\,\,\mathtt{wheeled(X).}\quad
    \mathtt{robot(X)}\leftarrow\,\,\mathtt{humanoid(X).} \quad
    \mathtt{ab(d}_{clmbstair}\mathtt{(X))}\leftarrow\,\,\mathtt{humanoid(X).}\\
    &\mathtt{wheeled(peoplebot).}\quad \mathtt{humanoid(nao).}\quad
    \textrm{\% Specific data} \vspace*{-0.4em}
  \end{align*}
}where {\small $\mathtt{ab(d(X))}$} implies ``X is abnormal with
respect to d''.  The result of inference in this example is:
$\lnot\mathtt{clmbstair(peoplebot)}$ \emph{without making any claims
  about} $\mathtt{nao}$, i.e., it is unknown if humanoid robot
$\mathtt{nao}$ can climb stairs or not.  These capabilities are
important for real-world human-robot collaboration. Inconsistencies
caused by incorrect information being added to the KB are identified
and corrected by processing sensor inputs or by posing queries to
humans.  The architecture (currently) only uses the inference
capabilities of ASP to better explore the merging of qualitative and
quantitative beliefs; future work will also consider ASP-based
planning.

\vspace{-1em}
\subsection{Uncertainty modeling with POMDP}
\begin{figure}[tcb]
  \begin{center}
    \includegraphics[width=0.7\textwidth]{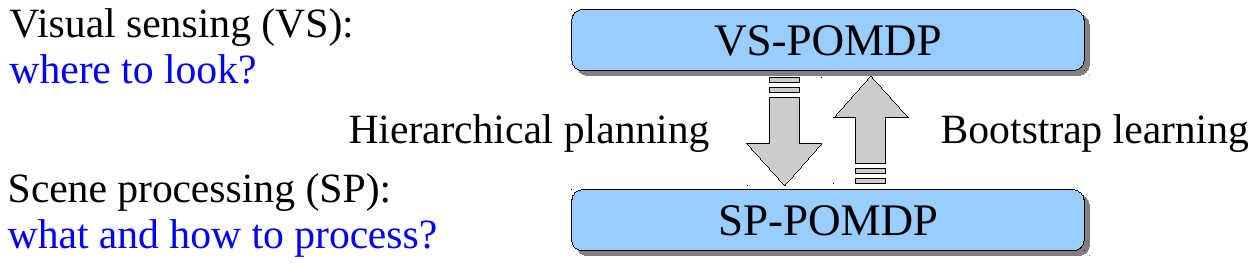}
    \vspace*{-1em}
    \caption{Hierarchical POMDPs for visual processing.}
    \label{fig:pomdps}
    \vspace*{-2em}
  \end{center}
\end{figure}

\vspace{-.5em}
\label{sec:pomdp}
To localize target objects, the robot has to move and analyze images
of different scenes. The uncertainty in sensor input processing and
navigation is modeled using hierarchical POMDPs as shown in
Figure~\ref{fig:pomdps}.  For a target object, the 3D area (e.g.,
multiple rooms) is modeled as a discrete 2D \emph{occupancy grid},
each grid storing the probability of target existence. The visual
sensing (VS)-POMDP plans an action sequence that maximizes information
gain by analyzing a suitable sequence of scenes.  For each scene, the
scene processing (SP)-POMDP plans the processing of specific regions
of images of the scene using relevant algorithms.  This formulation
uses our prior work on automatic belief propagation and model creation
in hierarchical POMDPs~\cite{zhang:aamas12}; some key features of the
hierarchy are described briefly.
Consider the VS-POMDP tuple $\langle S, A, T, Z, O, R\rangle$ for
locating an object in a grid with $N$ cells:
\begin{small_ind_s_itemize}
\item $S: \{s_{i}, \,\,i\in [1, N]\}$ is the state vector; $s_{i}$
  is the event that the target is in grid cell $i$.
\item $A: \{a_{i}, i \in [1, N]\}$ is the set of actions; $a_{i}$
  causes the robot to move and analyze cell $i$.
\item $T: S\times A\times S' \to [0, 1]$ is the state transition
  function.
\item $Z: \{\textrm{present, absent}\}$ is the observation set that
  indicates if the target is detected.
\item $O: S\times A\times Z \to [0, 1]$ is the observation
  function (see below).
\item $R: S\times A\times S' \to \mathbb{R}$ is the reward
  specification (see below).
\end{small_ind_s_itemize}
Since the state is not directly observable, the robot maintains a
\emph{belief state}, a probability distribution over the states.  In
the absence of prior knowledge, the belief state is a uniform
distribution. Information gain is maximized by computing a sequence of
actions that causes the belief distribution to converge to likely
target locations. The reward for executing action $a_t$ at time $t$ is
thus defined as the reduction in entropy between belief state
$B_{t-1}$ and the resultant belief state $B_{t}$.  The observation
function models the probability of each observation for each action
and state as a Gaussian distribution whose mean and variance depend on
the target location, cell being examined, sensor's field of view, and
distance to the object---robots learn this observation function
(Section~\ref{sec:expres}). A POMDP solver then computes a
\emph{policy} $\pi: B_t\mapsto a_{t+1}$ in the form of a matrix of
``weights'' that is used to probabilistically select an action for any
given belief state.

Executing an action using the VS policy causes the robot to move and
analyze a specific scene, automatically creating and solving the
corresponding SP-POMDP (with one or more levels). The robot uses the
SP policy to apply a sequence of visual processing algorithms on a
sequence of regions in images of the scene. The SP policy's terminal
action causes a VS belief update and subsequent action selection.  Our
hierarchical decomposition uses \emph{convolutional policies} and
learned observation functions to support \emph{automatic belief
  propagation} between levels of the hierarchy and \emph{real-time
  model and policy creation} in each level. Thus, robots
automatically, reliably and efficiently adapt visual processing and
navigation to the task at hand~\cite{zhang:aamas12}.

\vspace{-1em}
\subsection{Generating Prior Beliefs from Answer Sets}
\vspace{-.5em}
\label{sec:initDistributions}
This section describes the generation of prior beliefs from the ASP
KB. For indoor target localization, the robot computes prior belief of
existence of desired objects in rooms using: (a) knowledge of the
hierarchy of object classes and specific objects in the domain; and
(b) postulates that capture object co-occurrence relationships. We
illustrate this approach below for target localization and visual
processing; some postulates (and their representation) may need to be
revised for other sensors or in other domains.

\vspace{-1em}
\paragraph*{\bf Postulate 1:} 
Existence (non-existence) of objects of a primary class (in a room)
provides support for the existence (non-existence) of other objects of
this class (in the room). The level of support is proportional to
logarithm of the number of objects, inspired by \emph{Fechner's law}
(psychophysics) that states that subjective sensation is proportional
to logarithm of stimulus intensity: \vspace*{-.5em}
\begin{align*}
  perception = ln(stimulus)+ const
\end{align*}
This law is applicable to visual processing, the primary source of
information in this paper. For a target object, support for its
existence in a room is thus given by: \vspace*{-0.5em}
\begin{align}
  \label{eqn:perception}
  \psi_n=\begin{cases} 0 ~&if~~a_n=0 \\
    ln(a_n)+\xi ~&otherwise \end{cases}
\end{align}
where $a_n$ is the number of (known) objects of the primary class (of
the target) in the room, and $\xi=1$ corresponds to $const$ above.
Certain objects are exclusive, e.g., there is typically one fridge in
a kitchen. Such properties can be modeled by rules in the KB and by
including other postulates (e.g., see below).

\vspace{-1em}
\paragraph*{\bf Postulate 2:}
As the number of known subclasses of a class increases, the influence
exerted by the subclasses on each other decreases.  This computation
is performed recursively in the object hierarchy from each primary
class to the lowest common ancestor (LCA) of the primary class and
target object. Equation~\ref{eqn:perception} is thus modified as:
\vspace*{-0.5em}
\begin{align}
  \label{eqn:hyp2}
  \psi_n=\begin{cases} 0 ~&if~~a_n=0 \\
    \dfrac{ln(a_n)+\xi}{\prod_{h=1}^{H_n}W_h} ~&otherwise
  \end{cases}
\end{align}
where $H_n$ is the height of the LCA of the target objects and the
primary class under consideration. For a class node on the path from
the primary class to the LCA, $W_h$ is the number of siblings at
height $h$; $W_1=1$.

\vspace{-1em}
\paragraph*{\bf Postulate 3:}
Prior knowledge of existence of objects in different primary object
classes \emph{independently} provide support for the existence of a
specific object (in a room). This postulate is used to merge evidence
from different sources, and works well in practice. Prior belief of
existence of a target object in room $k$ is thus the summation of
evidence from the $N$ primary classes: \vspace*{-1.2em}
\begin{align}
  \label{eqn:dir_init}
  \alpha_k = \sum_{n=1}^{N}\psi_{n,k}
\end{align}
The prior belief of existence of the target in the set of rooms is
modeled as a Dirichlet distribution with parameter set
$\mathbf{\alpha}$, where $\alpha_k$ is computed (as above) using the
cardinality of the set of relevant answer set statements obtained
through ASP-based inference. The probability density function (pdf) of
this $K$-dimensional Dirichlet distribution is: \vspace*{-0.7em}
\begin{align}
  \label{eqn:kDimDir}
  \mathcal{D}(\mu|\alpha)=\frac{\Gamma(\alpha_0)}
  {\Gamma(\alpha_1)\cdots \Gamma(\alpha_K)}\prod^{K}_{k=1}
  \mu_k^{\alpha_k-1}
\end{align}
where $\Gamma$ is the Gamma function used for normalization;
$\mu_k\in[0,1]$ is a distribution of the target's existence over the
rooms; and $\alpha_0=\sum_{k=1}^{K}\alpha_k$.  The expectation of this
Dirichlet distribution, $\mathbb{E}(\mu_k)$, serves as the prior
belief distribution that is to be merged with the POMDP belief
distributions. If event $E_k$ represents the target's existence in
room $k$, the probability that the target is in room $k$ based on the
Dirichlet prior is: \vspace*{-.5em}
\begin{align}
  \label{eqn:dir_prior}
  p(E_k|\mathcal{D})=\mathbb{E}(\mu_k)=\alpha_k/\alpha_0
\end{align}
Although the Dirichlet distribution models the \emph{conditional}
probability of existence of the target in each room \emph{given} its
existence in the domain, it does not address the objective of learning
from positive and negative observations, thus reasoning about the
target's non-existence in a room or the domain. Towards this
objective, parameters $\mathbf{\alpha}$ also initialize Beta
distributions that model the existence of the target in each room:
\vspace*{-.4em}
\begin{align}
  \mathcal{B}(\phi_k|\alpha_k,
  \beta_k)=\frac{\Gamma(\alpha_k+\beta_k)}
  {\Gamma(\alpha_k)\Gamma(\beta_k)}\phi^{\alpha_k-1}(1-\phi^{\beta_k-1})
\end{align}
where $\Gamma$ function is used for normalization and $\beta_k$ is the
support for the non-existence of target in room $k$. The Beta
distribution's expectation for room $k$, $\mathbb{E}(\phi_k)$, is the
prior belief that the target exists in room $k$, and event $E$
represents the target's existence in the entire domain:
\vspace*{-0.5em}
\begin{align}
  \label{eqn:betaP}
  p(E_k|\mathcal{B})=\mathbb{E}(\phi_k)=\frac{\alpha_k}{\alpha_k+\beta_k}
\end{align}
The probability that the target does not exist in the domain can then
be derived, assuming $E_i$ and $E_j$ are independent events $\forall
i\ne j$: \vspace*{-0.5em}
\begin{align}
  \label{eqn:betaRooms}
  p(\neg E)=\prod_k{p(\neg
    E_k|\mathcal{B})}=\prod_k(1-{p(E_k|\mathcal{B}))} 
\end{align} \vspace*{-1.2em}

\noindent The Dirichlet distribution thus directly uses the information in the
answer sets to model conditional probabilities, while Beta distributions model
specific (marginal) probabilities that can be updated by positive and negative
observations (see below).

\vspace{-1em}
\subsection{Using Positive and Negative Observations}
\vspace{-0.5em}
\label{sec:nonexist}
Although it is a significant challenge to perform inference using lack of
evidence, negative observations can be used to identify eventualities
not modeled by the POMDPs, e.g., supporting early termination of tasks
that cannot be accomplished. For instance, not observing the target or
other related objects in multiple rooms can be used to reason about
the object's absence in the domain; this is not computed in the
standard POMDP model, and introducing a (special) terminal state in
the POMDPs invalidates the invariance properties exploited for
computational efficiency.

This section describes our approach to exploit all observations. Let
$D$ be the event that the target is detected; and $FoV$ the event that
target is in the robot's field of view.  The objective is to calculate
$p(\neg E|D)$ and $p(\neg E|\neg D)$, and thus $p(E|D)$ and $p(E|\neg
D)$, given the POMDP belief state $B$ and action $a$.  Towards this
objective, the distribution of target existence and non-existence (in
the domain) are updated along with the standard POMDP belief update.
The prior beliefs computed in Section~\ref{sec:initDistributions} are
(re)interpreted as beliefs conditioned on the existence (or
non-existence) of the target in the domain.

\vspace{-1em}
\paragraph*{\underline{\bf Negative Observations:}}
If the target is not detected, this observation should not change the
probability of target's existence outside the robot's field of view,
$p(\neg FoV|\neg D)$, which is the product of the probability that the
target exists in the domain, and the probability that the target
exists outside the robot's view given its existence in the
domain---the latter probability can be computed from the POMDP
beliefs.  The reduction in the probability of target's existence in
the field of view will be added to the probability of the target's
non-existence in the domain:
\vspace*{-0.5em}
\begin{align}
  \label{eqn:notDetected}
  \nonumber
  p(\neg E|\neg D) &=p(\neg E)+p(E)(p(FoV|E)-p'(FoV|E))\\
  &=p(\neg E)+p(E)(\sum_{s_i \in \Lambda(a)}{B(s_i)}-\sum_{s_i \in
    \Lambda(a)}{B'(s_i)}) 
\end{align}
\vspace*{-1em}

\noindent where $B(s_i)$ and $B'(s_i)$ are the POMDP beliefs of state $s_i$
before and after the Bayesian update using this observation, while
$\Lambda$, a function of action $a$, is the set of states that imply
that the target is within the robot's field of view.

\vspace{-1em}
\paragraph*{\underline{\bf Positive Observations:}} 
A positive observation should increase POMDP beliefs in the field of
view, and decrease beliefs outside this region and the probability of
target's non-existence in the domain. The posterior probability of
target's non-existence is computed based on the probabilities of the
target being detected given that it exists (or does not exist) in the
domain. The probability of the target being detected when it does not
exist is set to be a fixed probability of false-positive observations.
The probability of target being detected when it exists depends on
whether it is inside or outside the robot's field of view.  If the
target is in the field of view, the probability of a positive
observation is given by the POMDP observation functions; if it is
outside the field of view, this probability is the probability of
false-positives. We can then compute: \vspace*{-0.4em}
\begin{align}
  \label{eqn:ed}
  p(\neg E|D) &=\frac{p(D|\neg E)p(\neg E)}{p(D|E)p(E)+p(D|\neg
    E)p(\neg E)} 
\end{align}
The conditional probability of detecting the target given that it
exists (does not exist) is: \vspace*{-1.5em}
\begin{align}
  \label{eqn:de}
  p(D|E)&= p(D|E,FoV)p(FoV|E)+ p(D|E,\neg FoV)p(\neg FoV|E)
  \\\nonumber &= p(D|FoV)p(FoV|E) + p(D|\neg FoV)p(\neg FoV|E)
  \\\nonumber &= \sum_{s_i \in \Lambda(a)}{p(D|s_i,a)B(s_i)} +
  \epsilon\sum_{s_i
    \notin \Lambda(a)}{B(s_i)}\\
  P(D|\neg E) &= P(D|\neg E,FoV)P(FoV|\neg E)+P(D|\neg E,\neg
  FoV)P(\neg FoV|\neg E) \\\nonumber &= P(D|\neg E,\neg FoV) =
  \epsilon\sum_{s_i \notin \Lambda(a)}{B(s_i)}
\end{align}
where $\epsilon$ is the probability of false positives, i.e.,
detecting a target when it does not exist, and $p(D|s_i,a)$ is
obtained from the POMDP observation functions.

\vspace{-1em}
\subsection{Belief Merging and Information Acquisition}
\vspace{-0.5em} 
\label{sec:belief-merge}
The KB contains domain knowledge, including information that may not
be directly relevant to the current task. The POMDP belief summarizes
all observations directly relevant to the current task, but these
observations are obtained with varying levels of uncertainty. Our
prior work heuristically generated a belief distribution from answer
sets and (weighted) averaged it with POMDP
beliefs~\cite{zhang:icdl12}. In this paper, the use of Dirichlet
distribution to extract prior beliefs from answer sets supports
Bayesian merging: \vspace*{-0.4em}
\begin{align}
  \label{eqn:belief_bayesian}
  p'(E_k) = \frac{p(E_k|\mathcal{D})\cdot p(E_k)}
  {\sum_i{p(E_i|\mathcal{D})}\cdot p(E_i)}
\end{align}
where $p(E_k)$ is the probability that target is in room $k$ based on
POMDP beliefs.

Our architecture enables the robot to acquire initial domain knowledge
by mining online repositories and from minimal human input. This
incomplete (and possibly inconsistent) knowledge is revised using
sensor inputs and high-level human feedback. If a human (specific
humans are \emph{not} modeled) is nearby and the belief state entropy
is high (e.g., for the object being localized), the robot draws the
human's attention, followed by a query about a room's accessibility or
an object's existence in a room. If the entropy is low, the human is
ignored except for safe navigation.  The robot currently does not
model the uncertainty in human feedback; information provided by
humans is added to the KB, and any associated inconsistencies are
identified and resolved. Verbal human-robot interaction (HRI) is based
on simplistic templates such as:

\vspace{-0.7em}
{\small
\begin{verbatim}
Robot: Where is the [object]?  Human: In [room]./I do not know.
Robot: Is [room] accessible?   Human: Yes./No./I do not know.
\end{verbatim} 
}
\noindent
The robot may observe unforeseen changes in object configurations and
the domain, e.g., a door that was open may be closed. Such
observations can be added into the KB. Unlike our prior work, the KB
is thus augmented and revised continuously, adding and eliminating
areas for subsequent analysis based on the task at hand.

\vspace*{-1em}
\section{Experimental Results}
\vspace*{-0.5em}
\label{sec:expres}
The following hypotheses were evaluated experimentally: (H1) combining
ASP and POMDP improves target localization accuracy and time in
comparison with the individual algorithms; (H2) the entropy-based
strategy enables a robot to make best use of human feedback; and (H3)
using positive and negative observations helps robots identify tasks
suitable for early termination.  During the evaluation of hypotheses
H1 and H2, the approach for using positive and negative observations
(Section~\ref{sec:nonexist}) is not included.  Furthermore, our
approach for generating and merging ASP and POMDP beliefs is compared
with: (a) not using prior beliefs from the ASP KB; (b) using relative
trust factors to merge beliefs~\cite{zhang:icdl12}; and (c) using
weights drawn from the Dirichlet distribution---likely to assign
higher significance to prior beliefs extracted from the KB---to merge
beliefs.

Experiments included some trials on mobile robots and extensive
evaluation in simulation, using object models and observation
functions learned on physical robots to realistically simulate motion
and perception~\cite{Li:aamas13}. For instance, a simulated office
domain consists of four rooms connected by a surrounding hallway in a
$15\times 15$ grid. Fifty objects in ten primary classes (of office
electronics) were simulated, and some objects were randomly selected
in each trial as targets whose positions were unknown to the robot.
The robot revises the KB, including the basic object hierarchy mined
from repositories, during experimental trials. Each data point in the
results described below is the average of $5000$ simulated trials. In
each trial, the robot's location, target objects and locations of
objects are chosen randomly.  A trial ends when belief in a grid cell
exceeds a threshold (e.g., $0.80$). Some trials include a time limit
(see below).

\vspace{-1em}
\subsection{Simulation Experiments}
\begin{figure*}[tbc]
  \begin{center}
    \vspace{-1em} \hspace{-0.3in}
    \subfigure[0.5\textwidth][ASP-based localization.]{
      \includegraphics[width=0.53\textwidth]{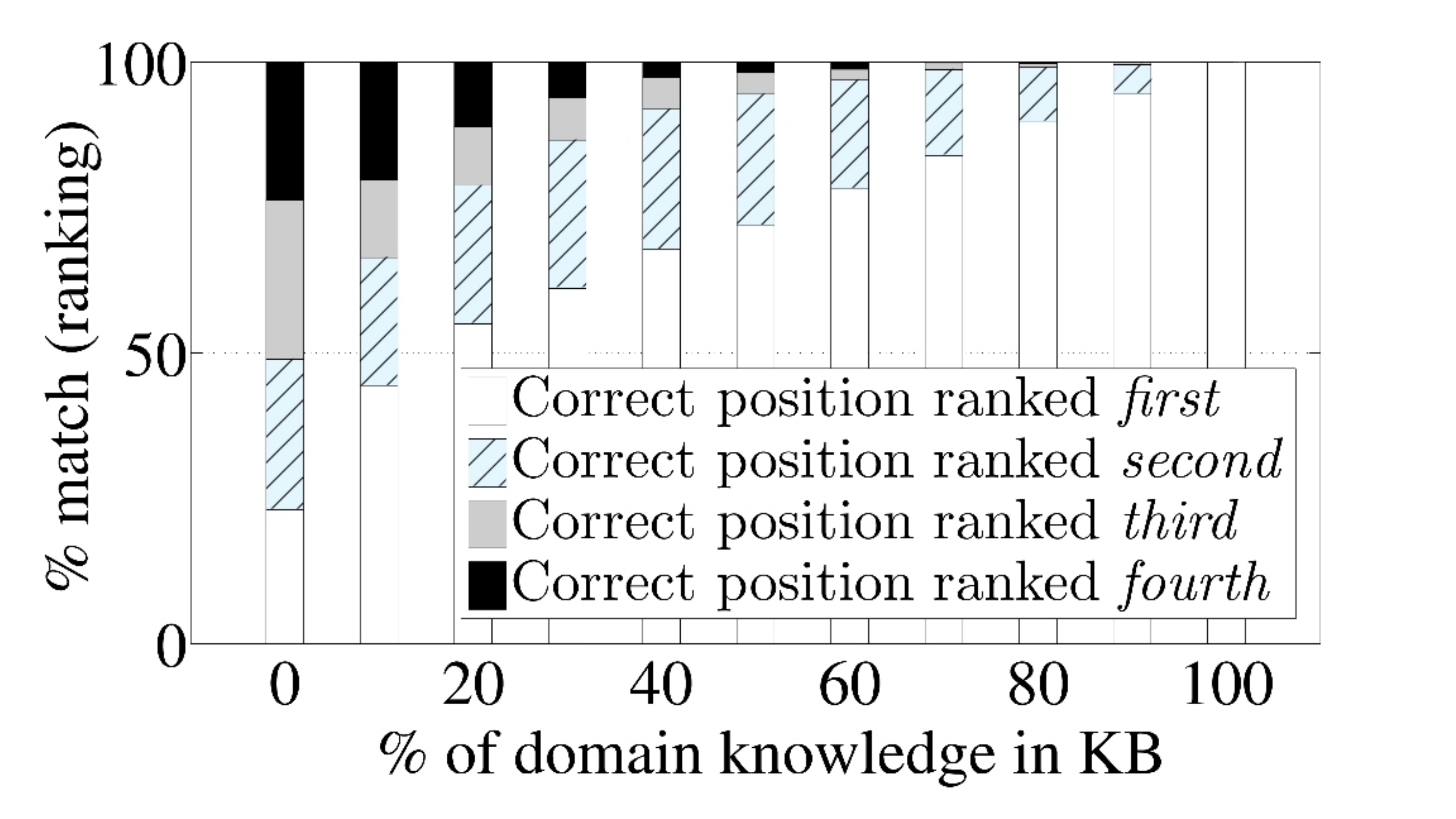}
      \label{fig:aspAccuracy}
    }\hspace{-0.2in}
    \subfigure[0.5\textwidth][ASP+POMDP localization.]{
      \includegraphics[width=0.53\textwidth]{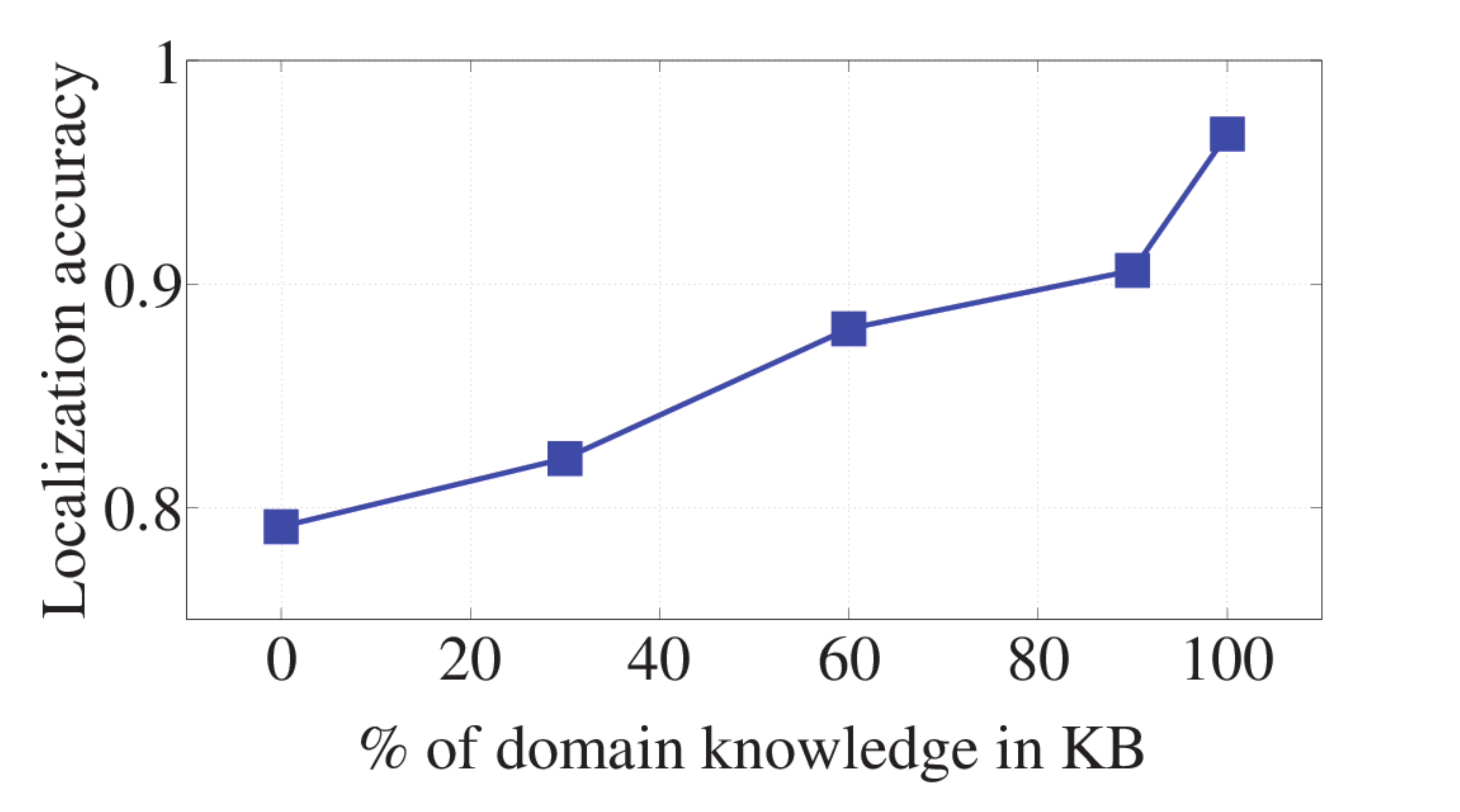}
      \label{fig:inference}
    }\hspace{-0.3in} \vspace{-1.3em}
    \caption{(a) Target localization accuracy using only ASP; (b)
      Target localization accuracy as a function of \% of domain
      knowledge encoded in the KB. Combining ASP and POMDPs
      significantly increases the accuracy in comparison with just
      using POMDPs (left-most point).}
    \vspace{-2em}
  \label{fig:asp_pomdp}
  \end{center}
\end{figure*}

\vspace{-0.5em}
\label{sec:simulate-test}
To evaluate hypothesis H1, we measured the target localization
accuracy and time when the robot used: (a) only ASP; (b) only POMDPs;
and (c) ASP and POMDPs. The robot first used domain knowledge and
Equations~\ref{eqn:perception}-\ref{eqn:kDimDir} to infer the target
objects' locations---ASP-based reasoning can state if a specific
object exists in a specific room but cannot provide the location of
the object in the room. Figure~\ref{fig:aspAccuracy} shows that when
the robot has all the domain knowledge (except the target), i.e.,
value$=100$ along x-axis, it can correctly infer the room containing
the object. The accuracy decreases when the domain knowledge
decreases, e.g., with $50\%$ of domain knowledge, the robot can
correctly identify the target's (room) location with $0.7$ accuracy.
However, even with $50\%$ domain knowledge, the correct room location
is in the top two choices in $95\%$ of the trials. One key expected
outcome of using ASP-based inference (and the associated beliefs) is
thus the significant reduction in target localization time.

For any target object, once ASP identifies the room(s) likely to
contain the object, POMDPs focus the robot's sensing and navigation to
determine the specific location of the object.
Figure~\ref{fig:inference} summarizes the results of these experiments
as a function of the amount of domain knowledge used to generate prior
beliefs. These trials have a time limit of $100$ units.  Trials
corresponding to $0$ on the x-axis represent the use of only POMDPs.
Combining prior beliefs extracted from answer sets with POMDP beliefs
significantly increases the target localization accuracy.  Although a
small amount of prior knowledge (selected randomly in simulation) can
cause the robot to waste time exploring irrelevant locations, reducing
accuracy when the time limit is exceeded and/or observations are
incorrect, the belief update helps the robot recover---accuracy is
higher if the time limit is relaxed. As the robot acquires and uses
more knowledge, localization accuracy steadily improves---with all
relevant domain knowledge (except the target), accuracy is $0.96$, and
errors correspond to trials with objects at the edge of cells.

\begin{figure*}[tbc]
  \begin{center}
    \hspace{-0.3in}
    \subfigure[0.5\textwidth][Comparing belief merging strategies.]{
      \includegraphics[width=0.53\textwidth]{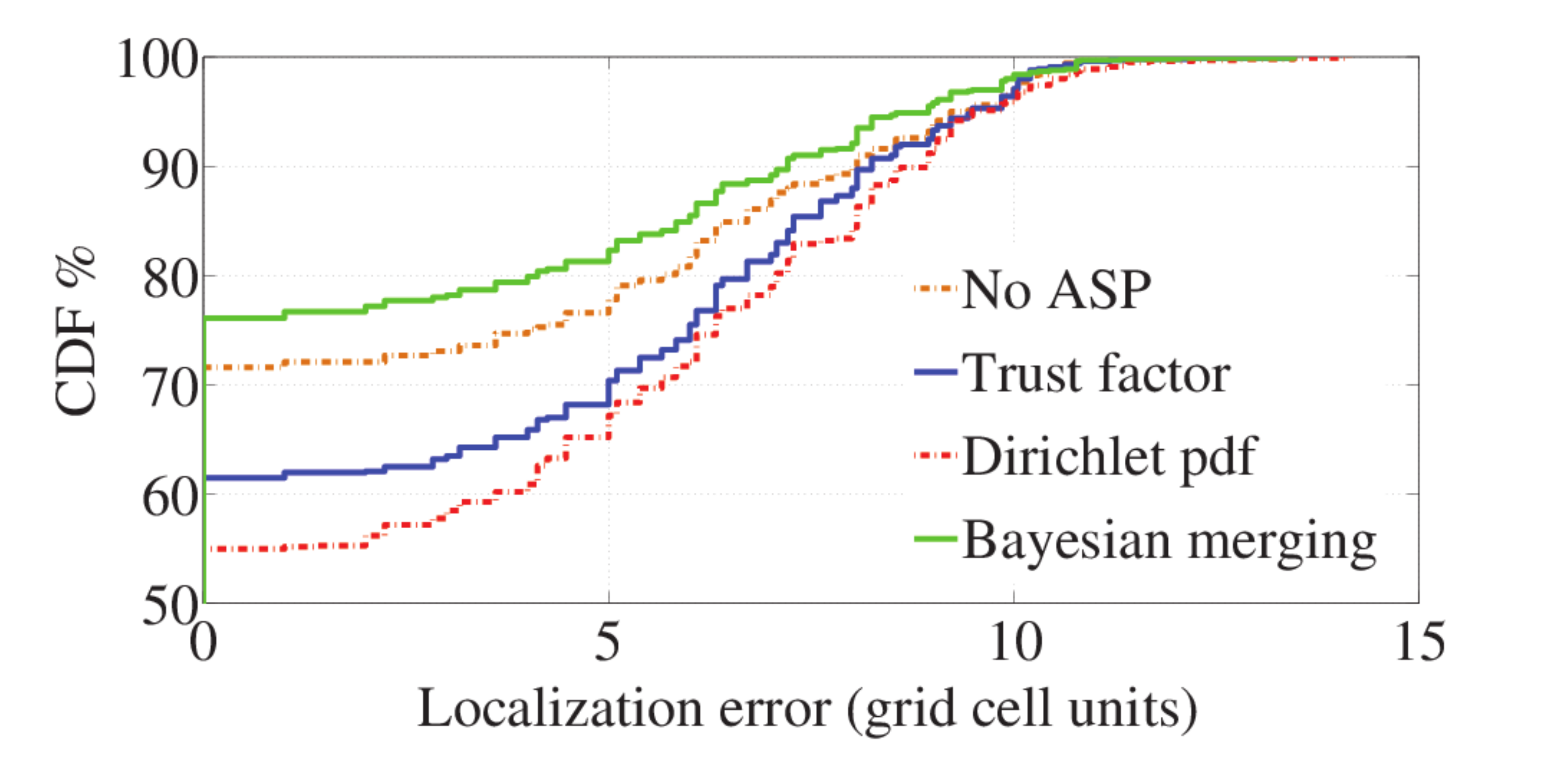}
      \label{fig:merging}
    }\hspace{-0.2in}
    \subfigure[0.5\textwidth][Human-robot interaction.]{
      \includegraphics[width=0.53\textwidth]{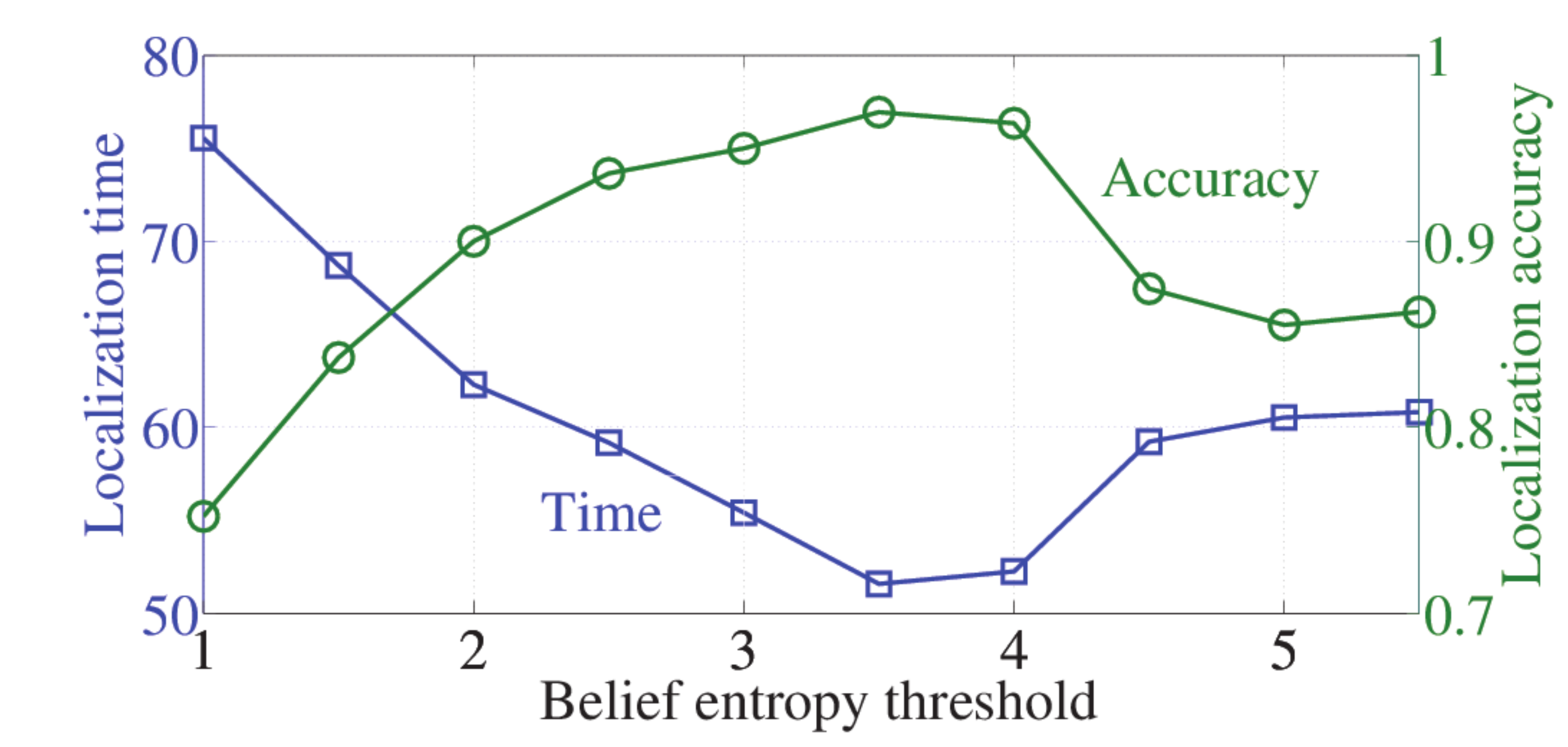}
      \label{fig:hri}
    }\hspace{-0.3in} \vspace{-1.3em}
    \caption{(a) Comparison of belief merging strategies: using
      Dirichlet distribution to model ASP-based prior beliefs, and
      Bayesian belief merging, results in lowest localization errors;
      (b) Entropy-based strategy to solicit human feedback: acquiring
      human feedback when needed increases localization accuracy while
      reducing localization time.}
    \vspace{-2.2em}
  \end{center}
\end{figure*}

Next, we evaluated our approach for generating and merging beliefs.
The KB is initialized with $20\%$ domain knowledge.  Periodically,
information about a few randomly chosen objects is added to the KB to
simulate learning from sensor inputs or human feedback.  Inference in
ASP produces new answer sets that are used to create prior beliefs to
be merged with the POMDP beliefs, thus guiding subsequent action
selection. As stated earlier, our approach is compared with three
other strategies, and Figure~\ref{fig:merging} summarizes the results.
The x-axis represents the localization error in units of grid cells in
the simulated domain, while the y-axis represents $\%$ of trials with
errors below a specific value. For instance, with our approach, more
than $80\%$ of the trials report an error of $\le 5$ units. These
(statistically significant) results indicate that our belief
generation and merging strategy results in much lower localization
errors than: not using ASP, our previous approach that used weighted
averaging (``trust factors'') to merge beliefs~\cite{zhang:icdl12},
and the Dirichlet-based weighting scheme that assigns weights to ASP
and POMDP beliefs based on the degree of correspondence with the
Dirichlet distribution. Similar results were observed with different
levels of (initial) knowledge---assigning undue importance to ASP or
POMDP beliefs can (in fact) hurt performance even when significant
domain knowledge is available.

To evaluate H2, i.e., the entropy-based strategy to solicit human
feedback, the robot is asked to localize target objects within a time
limit, e.g., $100$ time units. The robot uses ASP and POMDP beliefs
(with Bayesian belief merging), starting with a fixed amount of domain
knowledge in each trial. The robot can terminate the search if the
POMDP beliefs converge, e.g., one of the cells has a belief larger
than $0.8$. A human appears with $0.5$ probability after every action.
Soliciting human feedback is modeled as taking twice as much time as
(i.e., twice the cost of) a normal POMDP action. The (simulated)
human's response may be correct, not useful or even incorrect.
Figure~\ref{fig:hri} summarizes results of these experiments. When the
entropy threshold is too small, the robot will always ask questions
(when human is nearby), consuming a considerable amount of time in
acquiring information.  This hurts the robot's ability to accurately
localize objects within the available time. At the same time, when the
entropy threshold is too high, the robot rarely asks for feedback,
which also hurts localization accuracy. However, over a wide range of
entropy values that represent true need for feedback---performance not
very sensitive to choice of threshold---the robot localizes targets
with high accuracy while minimizing localization time.

\begin{figure*}[tbc]
  \begin{center}
    \hspace{-0.3in}
    \subfigure[0.5\textwidth][]{
      \includegraphics[width=0.53\textwidth]{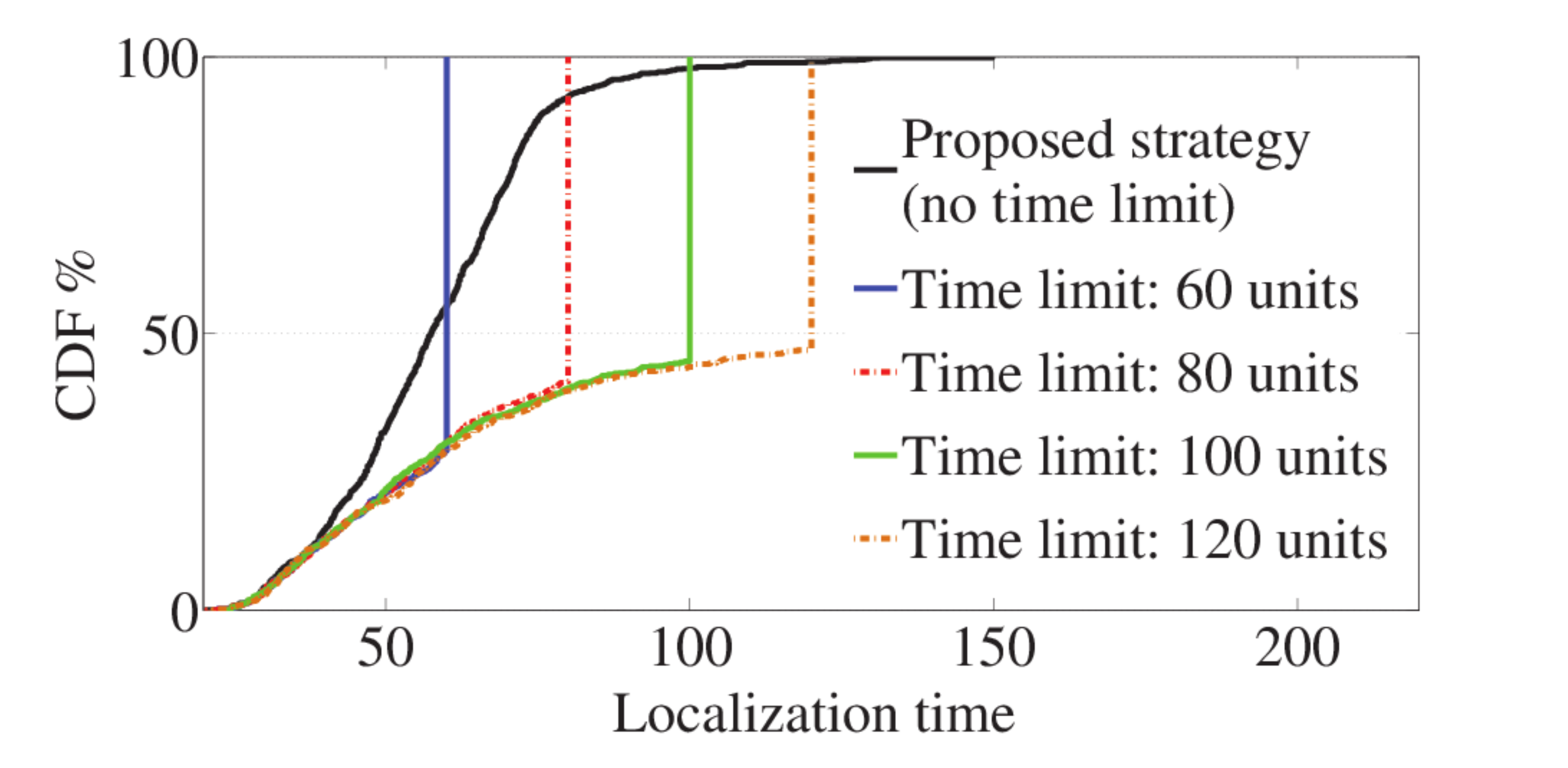}
      \label{fig:noTarget}
    }\hspace{-0.2in}
    \subfigure[0.5\textwidth][]{
      \includegraphics[width=0.53\textwidth]{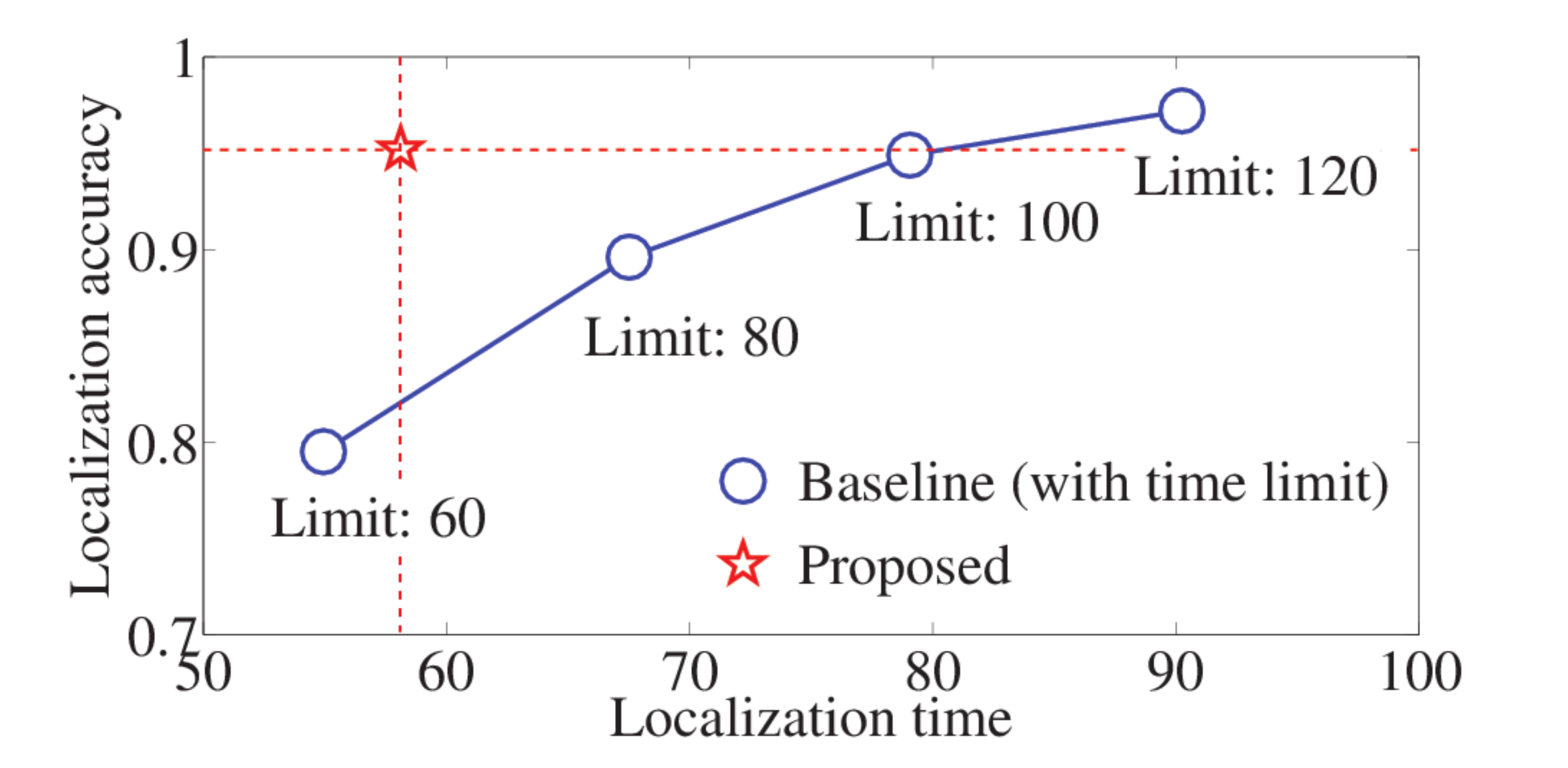}
      \label{fig:aveNoTarget}
    }\hspace{-0.3in} \vspace{-1.3em}
    \caption{(a) CDF of localization time: using positive and negative
      observations enables early termination of trials when the target
      does not exist in the domain; (b) Using positive and negative
      observations results in higher accuracy and lower localization
      time.}
    \label{fig:combinedNoTarget}
    \vspace{-2.2em}
  \end{center}
\end{figure*}

Experiments were then conducted to evaluate hypothesis H3, i.e., the
use of positive and negative observations. In each trial, the KB is
fixed, and the target is randomly selected to be present or absent. A
baseline policy (for comparison) is designed to use ASP and POMDP
beliefs (similar to experiments described above); this policy claims
absence of target if it cannot be found within the time limit.
Figure~\ref{fig:combinedNoTarget} summarizes these results.
Figure~\ref{fig:noTarget} shows that exploiting positive and negative
observations enables the robot to complete trials within $75$ time
units in $90\%$ of the trials; trials are mostly completed much before
the time limit especially when the target does not exist in the
domain.  Figure~\ref{fig:aveNoTarget} also shows that exploiting
positive and negative observations results in much higher target
localization accuracy while significantly lowering localization time;
if all observations are not exploited, comparable localization
accuracy is only obtained by allowing a much larger amount of time for
localization.

\vspace{-1em}
\subsection{Robot Experiments}
\begin{figure*}[tbc]
  \begin{center}
    \subfigure[0.41\textwidth][Learned domain map.] {
      \includegraphics[width=0.41\textwidth]{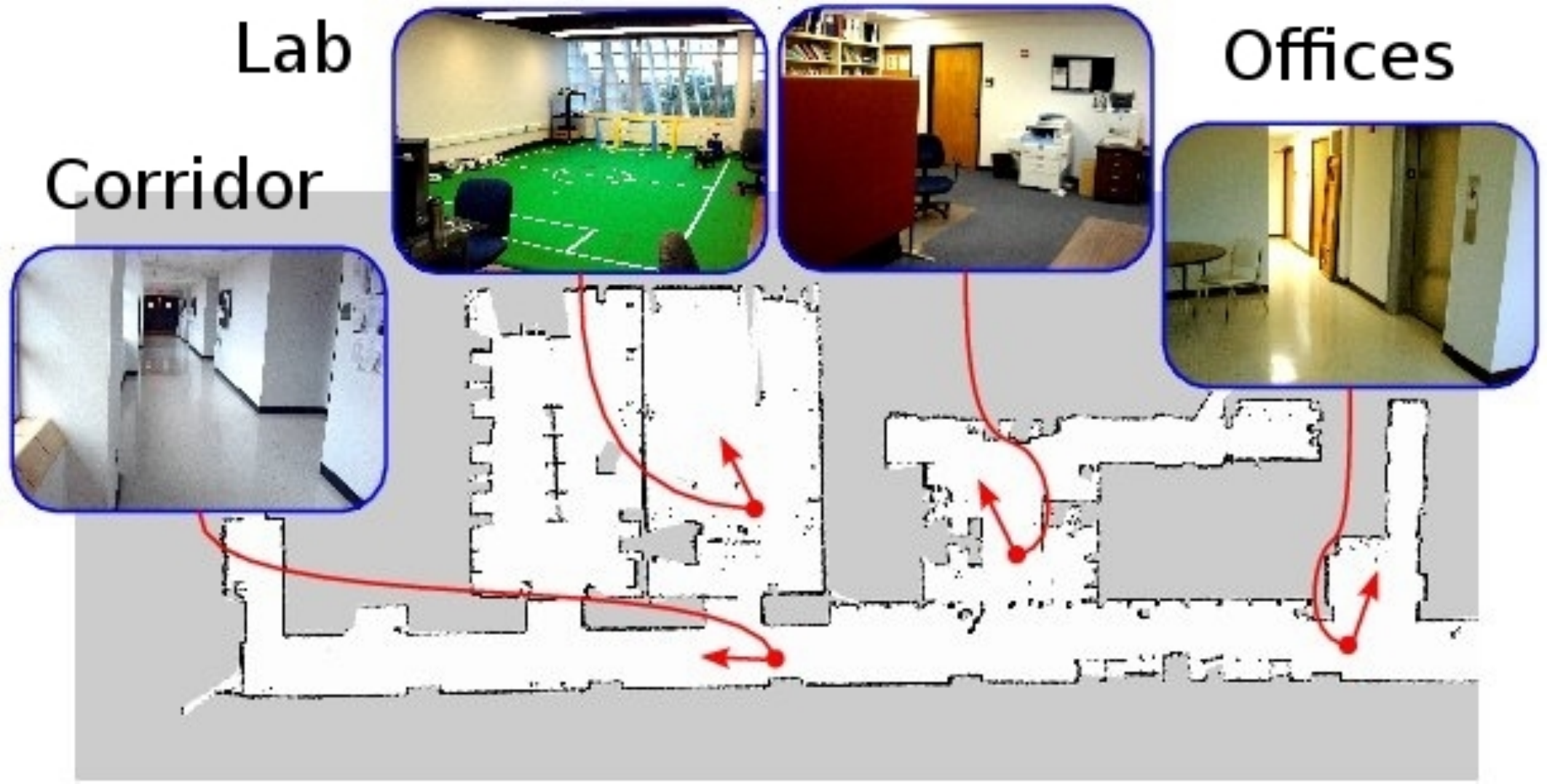}
      \label{fig:robot-sub1}
    } \hspace*{0.1in}
    \subfigure[0.41\textwidth][Target objects.] {
      \includegraphics[width=0.41\textwidth]{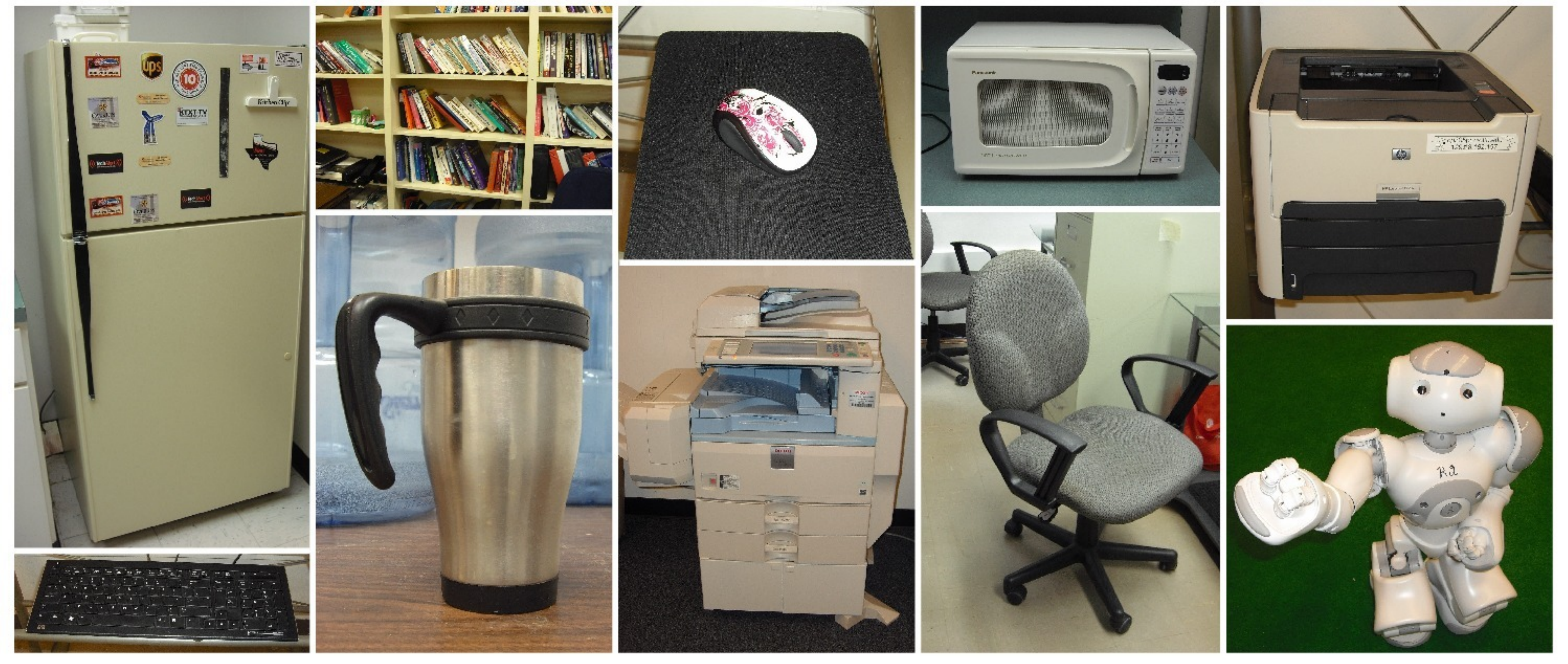}
      \label{fig:robot-sub2}
    }
  \end{center}
  \vspace{-1.75em}
  \label{fig:robot-exp}
  \caption{(a) Examples of target objects in experimental trials; and
    (b) Learned map of one floor with multiple labs, offices and
    corridors.}
  \vspace{-1em}
\end{figure*}

\vspace{-0.5em}
\label{sec:robot-test}
We evaluated the architecture on a wheeled robot deployed in an indoor
office domain. The robot learns (and revises) a domain map with
offices, corridors, labs and common areas, e.g.,
Figure~\ref{fig:robot-sub1}, and localizes the desired target
objects---Figure~\ref{fig:robot-sub2}. All algorithms were implemented
in the Robot Operating System (ROS). We compared our architecture with
two baseline strategies: (a) only POMDP beliefs; and (b) a heuristic
policy that makes greedy action choices.

The robot successfully localized target objects in all experimental
trials---there was no time limit. Since target localization times vary
depending on the initial positions of robot and targets, we report
target localization times of the baseline strategies as a factor of
the target localization time obtained using our architecture. Over
$30$ trials with each target object, the localization time with just
POMDP beliefs is $\approx 1.6$ times (averaged over all targets) that
with our architecture, while the factor is $\approx 2.4$ for the
heuristic policy. As observed in simulation trials, trusting ASP
beliefs a lot more than POMDP beliefs reduces localization
accuracy---just using ASP beliefs results in trials where the targets
are not found even after a long period of time.  Furthermore,
judicious use of human feedback enables the robot to further reduce
target localization time.

During the trials, the robot revises the KB and generates relevant
POMDP policies in real-time.  Consider a trial where the robot knows
that a refrigerator and a microwave exist in the ``kitchen'' and has
to localize a coffee maker.  The robot merges the learned (prior)
knowledge of object classes (in the answer sets) with POMDP beliefs to
obtain high initial belief that the coffee maker is in the kitchen. As
the robot moves to the kitchen, it meets a human but does not ask for
input because the belief entropy is not high. In the office outside
the kitchen, the robot detects an HP printer that had recently been
moved from the floor above, and the door to an instructor's office
that was closed recently. These facts, although not relevant to the
current task, revise the KB.  When the robot reaches the kitchen, it
processes images to localize the coffee maker.  If the robot now has
to enter the instructor's office or find the HP printer, it uses the
existing knowledge to automatically generate suitable initial belief
distributions, soliciting human
feedback appropriately.  Videos of some trials are available online:\\
\url{http://youtu.be/EvY_Jt-5BqM}, \url{http://youtu.be/DqsR2qDayGQ}

\vspace*{-0.7em}
\section{Conclusions}
\vspace*{-.5em}
\label{sec:conclusions}
This paper described an architecture that integrates the knowledge
representation and reasoning capabilities of ASP with the
probabilistic uncertainty modeling capabilities of hierarchical
POMDPs. Experimental results in the context of a mobile robot
localizing target objects in indoor (office) domains indicates that
the architecture enables robots to: (a) represent, revise and reason
with incomplete domain knowledge acquired from sensor inputs and human
feedback; (b) generate and merge ASP-based beliefs with POMDP beliefs
to tailor sensing and navigation to the task at hand; and (c) fully
utilize all observations for early termination of tasks that cannot be
accomplished. Future work will further explore the planning
capabilities of ASP, consider more complex application domains, and
investigate the tighter coupling of declarative programming and
probabilistic reasoning towards the long-term objective of reliable
and efficient human-robot collaboration in real-world application
domains.

\vspace{-1em}
\bibliographystyle{splncs}
\bibliography{references}

\begin{thebibliography}{10}

\bibitem{zhang:icdl12}
Zhang, S., Sridharan, M., Bao, F.S.:
\newblock {ASP+POMDP: Integrating Non-monotonic Logical Reasoning and
  Probabilistic Planning on Robots}.
\newblock In: International Conference on Development and Learning and
  Epigenetic Robotics. (November 7-9, 2012)

\bibitem{Gobelbecker:aaai11}
G\"obelbecker, M., Gretton, C., Dearden, R.:
\newblock {A Switching Planner for Combined Task and Observation Planning}.
\newblock In: National Conference on Artificial Intelligence (AAAI). (2011)

\bibitem{Pineau:RAS03}
Pineau, J., Montemerlo, M., Pollack, M., Roy, N., Thrun, S.:
\newblock {Towards Robotic Assistants in Nursing Homes: Challenges and
  Results}.
\newblock In: RAS Special Issue on Socially Interactive Robots. Volume~42.
  (2003)  271--281

\bibitem{Rosenthal:aaai11}
Rosenthal, S., Veloso, M., Dey, A.:
\newblock {Learning Accuracy and Availability of Humans who Help Mobile
  Robots}.
\newblock In: {National Conference on Artificial Intelligence}, San Francisco
  (2011)

\bibitem{zhang:aamas12}
Zhang, S., Sridharan, M.:
\newblock {Active Visual Sensing and Collaboration on Mobile Robots using
  Hierarchical POMDPs}.
\newblock In: Autonomous Agents and Multiagent Systems (AAMAS). (2012)

\bibitem{Baral:book03}
Baral, C.:
\newblock {Knowledge Representation, Reasoning and Declarative Problem
  Solving}.
\newblock Cambridge University Press (2003)

\bibitem{Gelfond:book08}
Gelfond, M.:
\newblock {Answer Sets}.
\newblock In {Frank van Harmelen and Vladimir Lifschitz and Bruce Porter}, ed.:
  {Handbook of Knowledge Representation}.
\newblock {Elsevier Science} ({2008})  {285--316}

\bibitem{Erdem:ISR2012}
Erdem, E., Aker, E., Patoglu, V.:
\newblock Answer set programming for collaborative housekeeping robotics:
  representation, reasoning, and execution.
\newblock Intelligent Service Robotics \textbf{5}(4) (2012)  275--291

\bibitem{Chen:JHRI2012}
Chen, X., Xie, J., Ji, J., Sui, Z.:
\newblock Toward open knowledge enabling for human-robot interaction.
\newblock Journal of Human-Robot Interaction \textbf{1}(2) (2012)  100--117

\bibitem{Hanheide:ijcai11}
Hanheide, M., Gretton, C., Dearden, R., Hawes, N., Wyatt, J., Pronobis, A.,
  Aydemir, A., Gobelbecker, M., Zender, H.:
\newblock {Exploiting Probabilistic Knowledge under Uncertain Sensing for
  Efficient Robot Behaviour}.
\newblock In: International Joint Conference on Artificial Intelligence
  (IJCAI), Barcelona, Spain (2011)

\bibitem{Kaelbling:iros11}
Kaelbling, L., Lozano-Perez, T.:
\newblock {Domain and Plan Representation for Task and Motion Planning in
  Uncertain Domains}.
\newblock In: {IROS Knowledge Representation for Autonomous Robots Workshop}.
  (2011)

\bibitem{richardson2006markov}
Richardson, M., Domingos, P.:
\newblock {Markov Logic Networks}.
\newblock Machine learning \textbf{62}(1) (2006)

\bibitem{Milch:bookchap07}
Milch, B., Marthi, B., Russell, S., Sontag, D., Ong, D.L., Kolobov, A.:
\newblock {BLOG: Probabilistic Models with Unknown Objects}.
\newblock In: {Statistical Relational Learning}.
\newblock {MIT Press} ({2006})

\bibitem{Li:aamas13}
Li, X., Sridharan, M., Meador, C.:
\newblock {Autonomous Learning of Visual Object Models on a Robot Using Context
  and Appearance Cues}.
\newblock In: {Autonomous Agents and Multiagent Systems (AAMAS)}. (2013)

\end{thebibliography}


\end{document}